\documentclass[journal]{IEEEtran}
\usepackage{cite}
\usepackage{amsmath,amssymb,amsfonts}
\usepackage{algorithmic}
\usepackage{graphicx}
\usepackage{textcomp}
\usepackage{booktabs}
\usepackage{booktabs}
\usepackage{siunitx}

\usepackage{stfloats}  

\begin{document}
\title{Physics-Inspired Gaussian Kolmogorov–Arnold Networks for X-ray Scatter Correction in Cone-Beam CT}
\author{Xu Jiang, Huiying Pan, Ligen Shi, Jianing Sun, Wenfeng Xu, Xing Zhao
\thanks{Xu Jiang, Jianing Sun, Wenfeng Xu, and Xing Zhao are with the School of Mathematical Sciences, Capital Normal University, Beijing 100048, China (e-mail: chianghsu97@gmail.com; 2220501015@cnu.edu.cn; willfore@163.com; zhaoxing\_1999@126.com).}
\thanks{Huiying Pan is with the School of Mathematics and Statistics, Nanjing University of Information Science and Technology, Nanjing, Jiangsu 210044, China (phy\_1995@126.com).}
\thanks{Ligen Shi is with the College of Computer Science (College of Software), Inner Mongolia University, Hohhot 010021, China (e-mail: ligenshi0826@gmail.com).}}

\maketitle

\begin{abstract}
Cone-beam CT (CBCT) employs a flat-panel detector to achieve three-dimensional imaging with high spatial resolution. 
However, CBCT is susceptible to scatter during data acquisition, which introduces CT value bias and reduced tissue contrast in the reconstructed images, ultimately degrading diagnostic accuracy.
To address this issue, we propose a deep learning-based scatter artifact correction method inspired by physical prior knowledge. 
Leveraging the fact that the observed point scatter probability density distribution exhibits rotational symmetry in the projection domain. The method uses Gaussian Radial Basis Functions (RBF) to model the point scatter function and embeds it into the Kolmogorov–Arnold Networks (KAN) layer, which provides efficient nonlinear mapping capabilities for learning high-dimensional scatter features. 
By incorporating the physical characteristics of the scattered photon distribution together with the complex function mapping capacity of KAN, the model improves its ability to accurately represent scatter.
The effectiveness of the method is validated through both synthetic and real-scan experiments.
Experimental results show that the model can effectively correct the scatter artifacts in the reconstructed images and is superior to the current methods in terms of quantitative metrics.
\end{abstract}


\section{Introduction}
\label{sec:introduction}
\IEEEPARstart{C}{omputed} tomography (CT) is an important diagnostic imaging tool\cite{hsieh2003computed} and has been widely used in the diagnosis various diseases\cite{pontone2022clinical, schambach2010application, weiss2019cone}. Cone-beam CT (CBCT) employs a flat-panel detector, providing high spatial resolution. However, the detector elements in flat-panel detectors are small and densely arranged, making it difficult to suppress X-ray scatter purely through hardware means\cite{gong2017physics}. The presence of scattered photons leads to shadow artifacts, reduces contrast, and lowers Hounsfield Unit (HU) values in reconstructed images, which significantly affects the imaging quality\cite{pivot2020scatter}. Siewerdsen and Jaffray\cite{siewerdsen2001cone} reported that when the system cone angle is large (e.g., a pelvis imaged with a cone angle of 6 degrees), the scatter-to-primary ratio (SPR) can even exceed 100\,\%. Therefore, an effective scatter correction method is crucial for maintaining the image quality and diagnostic value of CBCT.

At present, researchers have proposed various CBCT scatter correction methods, including hardware-based methods and software-based methods. Hardware-based methods aim to prevent scattered photons from reaching the detector. A simple method is to increase the distance between the object to be measured and the detector, which is called the air-gap method \cite{neitzel1992grids, persliden1997scatter}. However, this method narrows the scanning field of view and reducews the number of detected photons, hereby lowering the image signal-to-noise ratio (SNR). Another method is to place an anti-scatter grid \cite{neitzel1992grids, Chan1990Studies, chan1985performance} in front of the flat-panel detector. Although this method can partially suppress scatter, the grid also absorbs primary X-rays and increases patient radiation exposure. Moreover, because flat-panel detectors in CBCT consist of small and densely packed detector elements, it is difficult to design a compatible anti-scatter grid, often leading to moire pattern and grid line artifacts in the image. In addition, most hardware-based methods will require modifications to the hardware platform, which increases the system cost.

Software-based methods typically estimate the scatter signal distribution from scatter-contaminated images, and then correct the projection data\cite{trapp2022empirical}. These methods do not require hardware modifications and are relatively straightforward to implement. Among them, the Monte Carlo (MC) method\cite{2004Fast, 2009An} provides highly accurate scatter estimation. It estimates the scatter distribution based on the prior structure of the images and simulating the interaction between individual photons and the object being measured. However, due to the large number of photons, even with the use of GPU acceleration technology, the computational cost is still high and it is difficult to deploy clinically\cite{xu2025accelerated}. The scatter kernel superposition method (SKS)\cite{love1987scatter, 1999Efficient, 2008Scatter, 2010ImprovedSKS} uses convolution operations to replace the microscopic interaction between photons and matter, and approximates the scatter distribution as the convolution of the projection and the Gaussian kernel. The SKS method is significantly faster than MC and can achieve near real-time scatter estimation. However, it requires prior calibration of the convolution kernel parameters, and compared with MC, it is generally less accurate and struggles to model scatter distributions in complex anatomical structures with high precision. Maslowski et al. \cite{linearBoltzmann} designed a new software tool to estimate scatter in X-ray projection images by deterministically solving the linear Boltzmann transport equation. Niu et al. \cite{2024niuUBES} further proposed a semi-analytical method for unified scatter correction using an ultrafast Boltzmann equation solver. Although Boltzmann equation-based methods can achieve more than a hundred-fold acceleration compared to MCGPU, they still fall short of real-time performance.

Deep learning methods aim to learn the mapping between scatter-contaminated and scatter-free images during the training phase, enabling rapid suppression of scatter effects with inference completed in seconds. They can be roughly divided into three categories: image-domain methods\cite{DRCNN,ganScatter,2023An,zhangxueren2023flipswim,yang2025dual}, projection-domain methods\cite{hansen2018scatternet,2019Comparing,2019Real,2020Evaluation,pix2pixgan} and model-driven hybrid methods\cite{2021TMIBspline,iskender2022scatter,SKSCNN}. Image-domain methods directly learn the mapping function\cite{DRCNN} from the scatter-contaminated to the scatter-free images. For example, Zhang et al. \cite{zhangxueren2023flipswim} employed the U-shape network to capture the advantages of detailed textures and used the Swin Transformer to understand global features, thereby accurately extracting shallow and deep features. Image-domain methods do not require the original projection data and are simple to implement. 
However, in reconstructed images, scatter originating from a single projection point can propagate across the entire image. Moreover, scatter artifacts often resemble beam-hardening artifacts, making image-domain methods less reliable and less interpretable.Projection domain methods usually learn the distribution of scatter signals from scatter-contaminated projections and then remove the influence of scatter from the original data\cite{MARC-DSEnet,hansen2018scatternet}. Jiang et al. \cite{pix2pixgan} used the pix2pix GAN network model that combines the residual module and patch GAN to train multi-spectral projection data labels to improve the accuracy of scatter correction. While projection-domain methods operate directly on measurement data that contain richer information, the lack of explicit physical or model-based constraints may lead to over-smoothing of certain details, thereby introducing secondary artifacts in the reconstructed images.

Recently, model-driven hybrid approaches have emerged, wherein traditional scatter models are embedded within deep neural networks, thereby enabling improved CBCT image reconstruction. Zhuo et al.\cite{SKSCNN} combined SKS with a convolutional neural network and used CNN to learn the amplitude and width maps of Gaussian kernels, thereby effectively enhancing the performance of SKS. 

Kolmogorov–Arnold Networks (KAN) are a recently proposed neural architecture inspired by the Kolmogorov–Arnold representation theorem. Unlike conventional networks that rely on fixed activation functions, KAN introduce learnable non-linear functions on the edges (``weights''), thereby enhancing their capacity for non-linear representation learning. Motivated by this idea, physics-informed priors of X-ray scatter are embedded into the basis functions of KAN to enable efficient and accurate modeling of scatter distributions for CBCT correction.

\begin{figure}[h]
    \centerline{\includegraphics[width=0.4\textwidth]{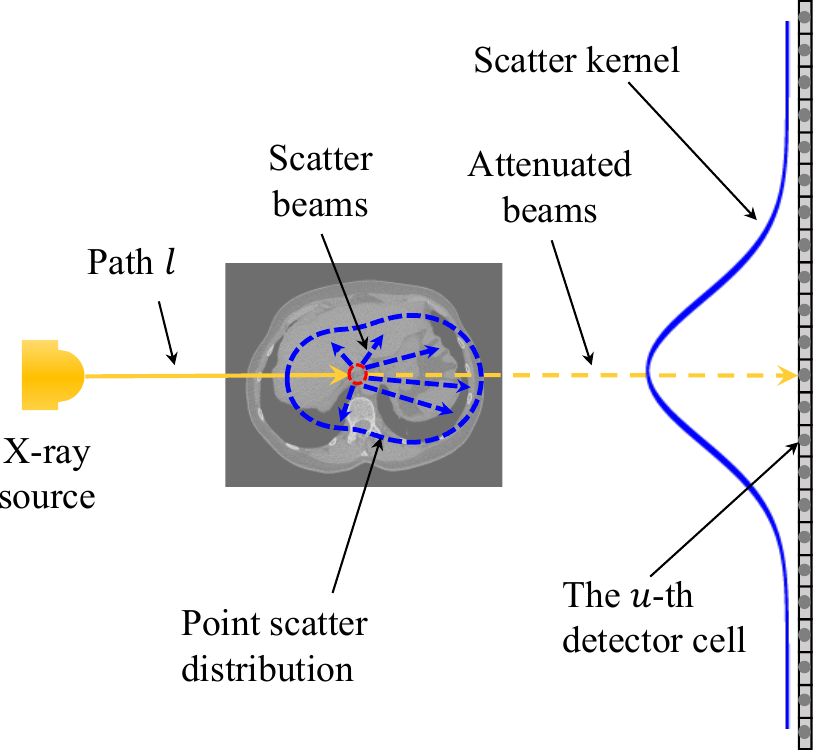}}
    \caption{Schematic diagram of two-dimensional scatter distribution characteristics in a three-dimensional CBCT system.}
    \label{fig:Fig.1.}
\end{figure}

Fig. \ref {fig:Fig.1.} is a two-dimensional schematic diagram of the point scatter distribution in a CBCT system. X-rays are emitted from the source, attenuated by the patient’s body, and reach the detector, during which scatter occurs. Scatter phenomenon is the result of the superposition of a large number of random processes occurring between photons and the object’s internal structures, and the scatter distribution is strongly dependent on the structure of object and system geometry.

We notice that Compton scatter dominates at typical medical X-ray energies, and the Compton scatter angle can be described by the Klein-Nishina formula\cite{fatima2023x}, the integration of which yields the effect cross section of Compton scatter in Eq. (\ref{eq:111}):
\begin{align}
    \sigma = 2\pi r_e^2 \bigg[ \frac{1 + \epsilon}{\epsilon^3} \left( \frac{2\epsilon(1 + \epsilon)}{1 + 2\epsilon} - \ln(1 + 2\epsilon) \right) \nonumber
    \\
    + \frac{\ln(1 + 2\epsilon)}{2\epsilon} - \frac{1 + 3\epsilon}{(1 + 2\epsilon)^2} \bigg],
    \label{eq:111}
\end{align}
where $\sigma$ denotes the point scatter probability distribution corresponding to the blue curve in Fig. \ref {fig:Fig.1.}, $r_e$ is classical electron radius, and $\epsilon$ represents the scatter angle, defined as the angle between the scattered and incident X-ray photons. 

Using the point scatter probability density function, the scatter probability distribution on the projection domain can be calculated, and their superposition yields the final scatter intensity distribution. From the detector’s perspective, the probability distribution of single point scatter event is centered on the transmitted ray and decays gradually in all directions. This distribution is rotationally symmetric and can be effectively approximated by Gaussian radial basis functions. Based on the expressive power of Gaussian radial basis functions, the KAN is used to model the complex phenomena of CBCT scatter by replacing the traditional fixed activation function with a learnable nonlinear activation function for efficient feature extraction. Therefore, this article proposes a projection domain scatter correction method. The contributions are as follows:
\begin{itemize}
    \item Leveraging the physical distribution characteristics of scattered signals in the projection domain, Gaussian radial basis functions are employed for modeling and feature extraction, thereby incorporating physics-informed prior constraints into the network.
    \item Learnable Gaussian RBF are embedded into the KAN layer, combined with a U-shape backbone to extract multi-scale features, improving both computational efficiency and model reliability.
    \item The proposed network is designed to learn scatter distributions. Owing to the low-frequency nature of scatter signals, down-sampling of the input projection-domain data effectively reduces the number of trainable parameters, thereby lightening the network architecture without imposing restrictions on projection size.
\end{itemize}

The remainder of this paper is organized as follows. Section II outlines the proposed methodology. Section III describes the experimental setup. Section IV presents evaluations on experiments.

\begin{figure*}[!t]
\centerline{\includegraphics[width=0.9\textwidth]{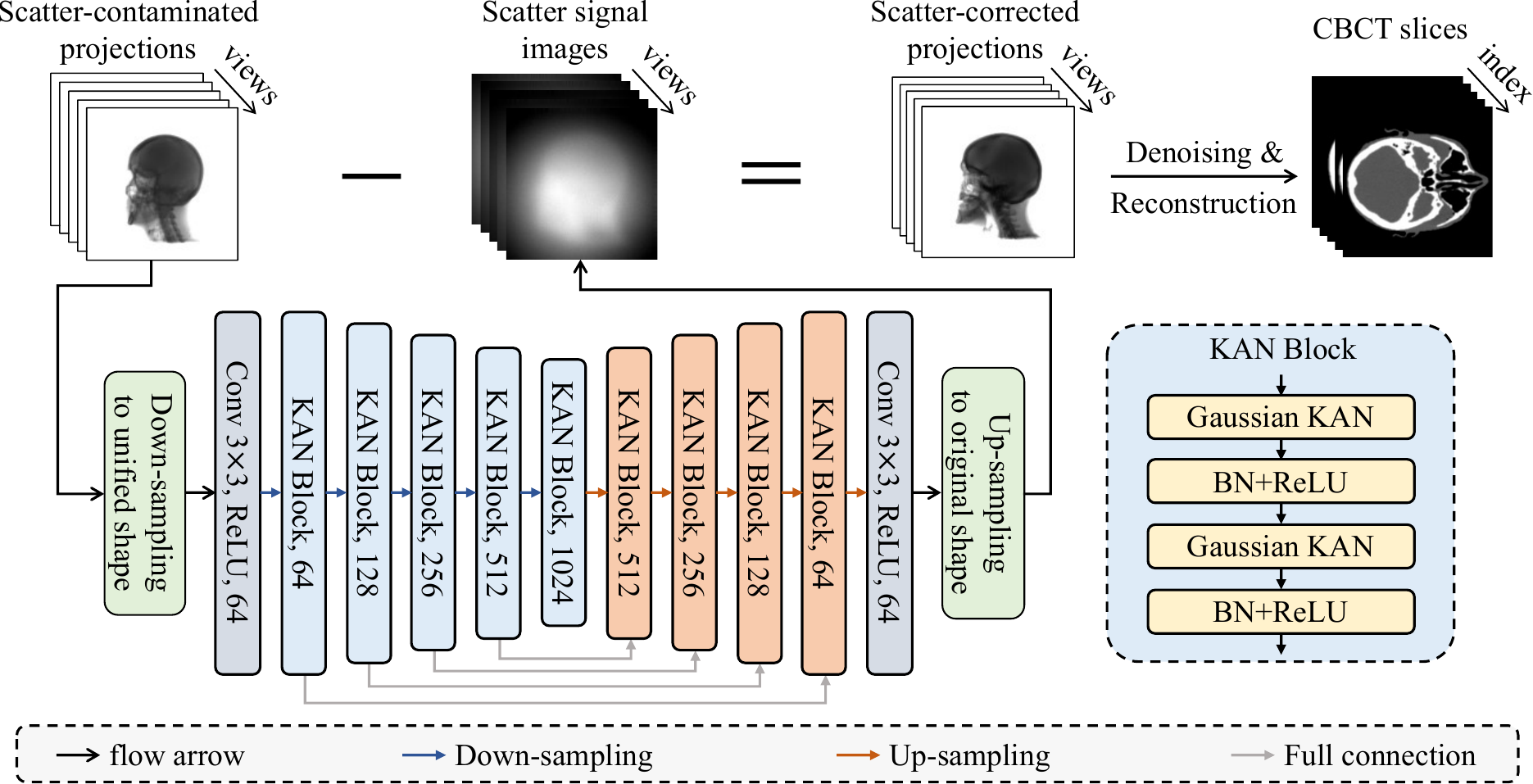}}
\caption{The overall architecture of the proposed method.The dotted blue part is the encoding part, and the dotted pink part is the decoding part.}
\label{fig:Fig.2.}
\end{figure*}

\section{Method}

In this section, we first present CBCT scatter correction model, followed by a description of the network architecture. The main framework of the processed method is shown in Fig. \ref{fig:Fig.2.}.

\subsection{CBCT Scatter Correction Model}

Under ideal conditions, the X-ray attenuation model in CBCT system follows Beer Lambert's law, and its integral form can be expressed as:

\begin{equation}
    I_p(u, v) = \int^{E_{\text{max}}}_{E_{\text{min}}} I_0(u, v, E) \, e^{-\int_{l} \mu(x, E) dx} dE,
    \label{eq:1}
\end{equation}
where $I_0 \in \mathbb{R}^{w \times h}$ denotes the incident polychromatic flat-field projection at energy $E$, $(u,v)$ are detector indices in a $w \times h$ array, $\mathbf{\mu}$ is the linear attenuation coefficient at spatial location $x$ and the energy $E$, and $I_p \in \mathbb{R}^{w \times h}$ is the primary (transmitted) projection. 

While Eq.~\eqref{eq:1} describes the ideal exponential attenuation of primary photons, in practical CBCT imaging photon–matter interactions are not limited to simple attenuation but also give rise to scatter processes such as Compton, Rayleigh, and multiple scatter. Accordingly, the scatter-contaminated projections $I_m \in \mathbb{R}^{w \times h}$ (as shown in Fig. \ref{fig:Fig.2.}) measured at the detector are commonly expressed as the superposition of the primary photons $I_p$ and the scatter signal images $I_s \in \mathbb{R}^{w \times h}$,

\begin{equation}
    I_m(u, v) = I_p(u, v) + I_s(u, v).
    \label{eq:2}
\end{equation}
Note that this model does not consider factors such as X-ray fluence gain, detector response, or noise. Therefore, the key to obtaining the ideal primary photon signal $I_p$ lies in accurately estimating the scatter signal $I_s$.

We propose a model-driven hybrid scatter estimation network $\mathcal{N}$ that maps the scatter-contaminated projections $I_m$ to the scatter signal images $I_{s}$ (as shown in Fig. \ref{fig:Fig.2.}). This mapping is expressed as $\tilde{I_{s}} = \mathcal{N}(I_m, \theta)$, where $\tilde{I_{s}}$ is the estimated scatter signal, and $\theta$ represents the learnable network parameters. Thus, the scatter correction problem is transformed into:

\begin{equation}
    \theta^{\ast} = \arg\min_{\theta} \frac{1}{d} \sum_{i=1}^{d} \left\| \mathcal{N} \left( \frac{I_m}{I_{0}}, \theta \right) -  \frac{I_{s}}{I_m}   \right\|,
    \label{eq:22}
\end{equation}
where $d$ is the number of projection views. $\frac{I_m}{I_{0}}$ represents the normalized projection to mitigate tube current effects. Due to the influence of scatter and noise, the actual measured value $I_m$ may exceed the incident intensity $I_{0}$, data truncation is applied to the regions where $\frac{I_m}{I_0}$ greater than $1$. Unlike general normalization, the target is $\tfrac{I_s}{I_m}$ rather than $\tfrac{I_s}{I_0}$. This choice reflects the fact that although $I_s \ll I_0$, regions with strong attenuation exhibit high SPR, making $\tfrac{I_s}{I_m}$ a better descriptor of the scatter distribution.

In a supervised learning framework, the optimal network parameters $\theta^{\ast}$ are obtained. During the correction process, the trained network takes the scatter-contaminated data $I$ as input and outputs the predicted scatter signal $\tilde{I}_{s}$,

\begin{equation}
    \tilde{I}_{s} = \mathcal{N} \left(\frac{I_m}{I_{0}}, \theta^{\ast} \right),
\end{equation}
Subtracting the predicted scatter signal from the original signal, scatter-corrected projections $\tilde{I}_{p}$ can be obtained. As scatter correction is a decoupled process, noise amplification may occur, making denoising essential. The denoising operator is denoted as $\mathcal{D}$, which typically employs median or mean filtering,

\begin{equation}
    \tilde{I}_{p} = \mathcal{D}(I_m - \tilde{I}_{s}).
\end{equation}
The photon-domain data is then converted to projection-domain data. Finally, the scatter-corrected projections are reconstructed to obtain a scatter-free image.

\begin{equation}
    g = \mathcal{R} \left (-ln \left(\frac{\tilde{I}_{p}}{I_{0}} \right) \right).
\end{equation}
Here $\mathcal{R}$ denotes the reconstruction operator, such as FDK algorithm or other reconstruction technology. At this point, the final scatter-corrected reconstruction is denoted as $g$.

\subsection{Kolmogorov-Arnold Network}


Inspired by Kolmogorov-Arnold representation theorem, Liu et al\cite{liu2024kan20kolmogorovarnoldnetworks} defined a generalized Kolmogorov-Arnold layer to learn univariate functions on edge, in the form of activation function. The KAN layer can be written as:

\begin{equation}
\begin{aligned}
    f(\mathbf{x}) = \Phi \circ \mathbf{x} &= 
    \left[ \sum_{i=1}^{d_{in}} \phi_{i,1}(x_i) \quad \cdots \quad \sum_{i=1}^{d_{in}} \phi_{i,d_{out}}(x_i) \right], \\
    \Phi &= 
    \begin{bmatrix}
    \phi_{1,1}(\cdot) & \cdots & \phi_{1,d_{in}}(\cdot) \\
    \vdots & \ddots & \vdots \\
    \phi_{d_{out},1}(\cdot) & \cdots & \phi_{d_{out},d_{in}}(\cdot)
    \end{bmatrix}.
\end{aligned}
\label{eq-KAN-layer}
\end{equation}
where $\Phi$ represents learnable parameters, it uses a linear combination of SiLU activation and a basis function $\operatorname{bf}(x)$,

\begin{equation}
\begin{aligned}
    \phi(x) = w_{1}\operatorname{SiLU}(x) + w_{2} \operatorname{bf}(x).
\end{aligned}
\label{eq-kan-comb}
\end{equation}
In most implementations, The basis function $\operatorname{bf}(x)$ in $\phi(x)$ uses a B-spline function. Here, motivated by the prior knowledge that Klein-Nishina formula exhibit symmetry in the projection domain, we adopt the Gaussian RBF as the basis function. In addition, B-splines are difficult to parallelize on GPUs\cite{yang2025kolmogorovarnold}, leading to low computational efficiency and limited scalability to high-dimensional spaces,
\begin{equation}
\begin{aligned}
    \operatorname{bf}(x) =\sum_{j=1}^{c}  w_{j}\exp\left(-\frac{(x - \mu_j)^2}{\sigma^2}\right).
\end{aligned}
\label{eq-kan-guassian}
\end{equation}
Here, $c$ denotes the number of channels, $\mu_j$ is the center, $\sigma$ is the scale parameter, and $w_j$ represents the trainable weight. For computational convenience, $c$, $\mu_j$, and $\sigma$ are fixed parameters. Geometrically, this formulation uses $c$ Gaussian kernels with identical scales but different centers to fit the signal via a weighted combination.

Unlike traditional MLPs and CNNs that utilize fixed activation functions for feature extraction, KAN are characterized by their ability to replace conventional fixed linear weights with learnable univariate functions, thereby enhancing the network's representational capacity.

\begin{figure}[t]  
    \centering
    \includegraphics[width=0.4\textwidth]{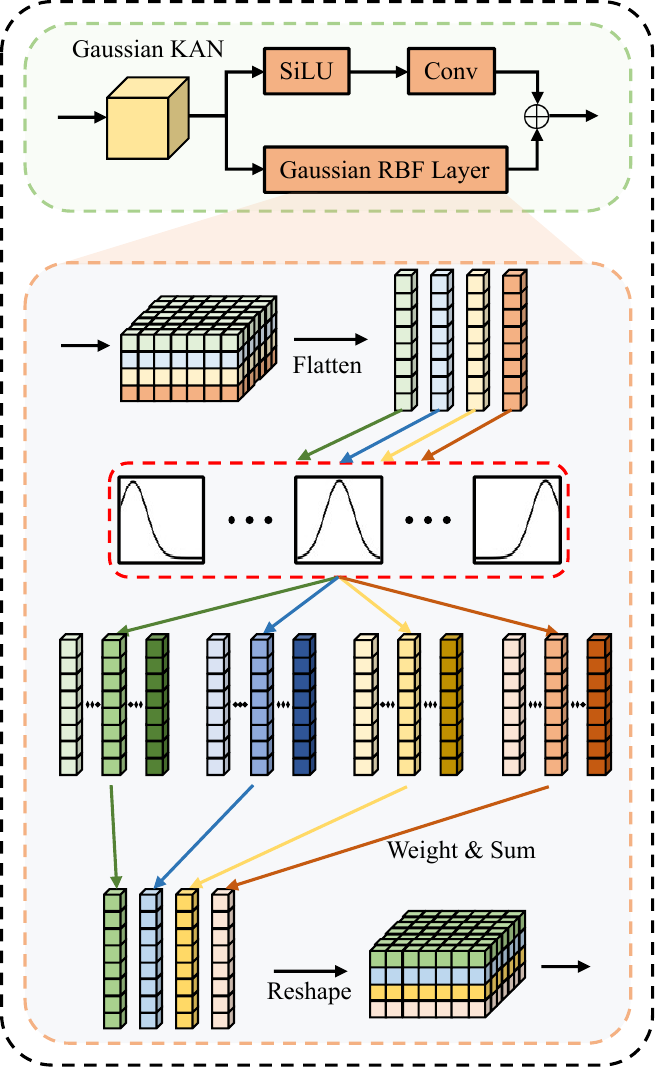} 
    \caption{The overall architecture of Gaussian KAN layer.} 
    \label{fig:Fig.3.} 
\end{figure}

\subsection{Network Architecture}

A U-shape architecture based on Gaussian RBF KAN is proposed for scatter correction. The encoder–decoder structure of the U-shape backbone effectively captures image features and has become a popular architecture in the field of medical imaging \cite{li2024ukanmakesstrongbackbone}. It leverages down-sampling and up-sampling to extract features at multiple scales, while skip connections are used to enhance feature representation.

Instead of traditional CNN modules for feature extraction, the proposed network employs KAN layers, as illustrated in Fig.~\ref{fig:Fig.3.}. The extracted features are divided into two parts.

\begin{align}
\mathcal{F}^{l+\frac{1}{2}}_{p1} &= \operatorname{Conv}(\operatorname{SiLU}(\mathcal{F}^{l})), \\
\mathcal{F}^{l+\frac{1}{2}}_{p2} &= \operatorname{GaussRBF}(\mathcal{F}^{l}), \\
\mathcal{F}^{l+1} &= \mathcal{F}^{l+\frac{1}{2}}_{p1} + \mathcal{F}^{l+\frac{1}{2}}_{p2}.
\end{align}

For the first part, denoted as $ \mathcal{F}^{l+\frac{1}{2}}_{p1} $, the operation is similar to conventional convolutional feature extraction. The second part, $ \mathcal{F}^{l+\frac{1}{2}}_{p2} $, as illustrated in Fig.~\ref{fig:Fig.3.}, involves the following steps: the input feature maps are first unfolded along the channel dimension into column vectors. For each column vector, a set of Gaussian kernels centered at different positions is applied to project the features into a higher-dimensional space. Then, a weighted summation is performed using learned weights, completing the RBF mapping process. Finally, the transformed features are reshaped back to the original size.

\section{Experiments}
\subsection{Data Description}

In practice, obtaining training datasets with accurate scatter information is time-consuming and labor-intensive. MC-GPU\cite{2009Accelerating}, an open-source software, was used to simulate paired datasets consisting of scatter-free and scatter-contaminated projection data. The input CBCT datasets were derived from the HNSCC-3DCT-RT\cite{TCIAclark2013cancer}, which is publicly available through The Cancer Imaging Archive (TCIA). First, the head region was segmented from the CBCT images in the dataset. Then, threshold-based segmentation was applied to classify the voxels into three material types: air, soft tissue, and bone. Subsequently, the corresponding materials were mapped to their respective mass densities based on the PENELOPE 2006 database. Finally, The processed data were then used as inputs to MC-GPU to generate paired projection dataset.

For each view, the projection was simulated using an initial total photon count of $1.25 \times 10^{10}$. To reduce beam hardening effects, a $1$\,mm thick copper (Cu) filter was placed in front of the X-ray source for spectral pre-filtration. The X-ray spectrum at 120\,kVp was generated using the Spectrum GUI\cite{SpectrumGUI}. The flat-panel detector consisted of $512 \times 512$ pixels, with each pixel measuring $0.8 \times 0.8\,\mathrm{mm}^2$. A full circular scan consisted of 360 sampling angles. We generated dataset using 10 head phantoms for network training and one additional head dataset for validation. As a result, the training dataset included 3600 paired projection samples, and 360 projections were used for validation.

\subsection{Experiments Details}

In this study, we employed Python and the PyTorch libraries to implement the framework and perform experiments on a server equipped with Gold 6248R 3.00Hz CPU, 256 GB RAM, and an NVIDIA 4090 24 GB GPU. The optimizer was Adam with $\beta_{1}=0.9$, $\beta_{2}=0.999$ and a weight decay of $10^{-4}$.The number of total epochs was $100$. The initial learning rate was set to $10^{-5}$, and decayed at $3000$ iterations respectively by multiplying $0.5$. The batch size was set to $1$. The validation experiments and the real-scan experiments shared the same training parameters.

\subsection{Comparison With Other Methods}

In order to verify the effectiveness of our approach, we conducted comparisons against several representative projection-domain methods, including: SKS method\cite{2010ImprovedSKS}, DSE-Net method\cite{MARC-DSEnet}, UNet-ViT method, Pix2pix GAN method\cite{pix2pixgan}.

\section{Results}

\begin{figure*}[t]  
    \centering
    \includegraphics[width=0.95\textwidth]{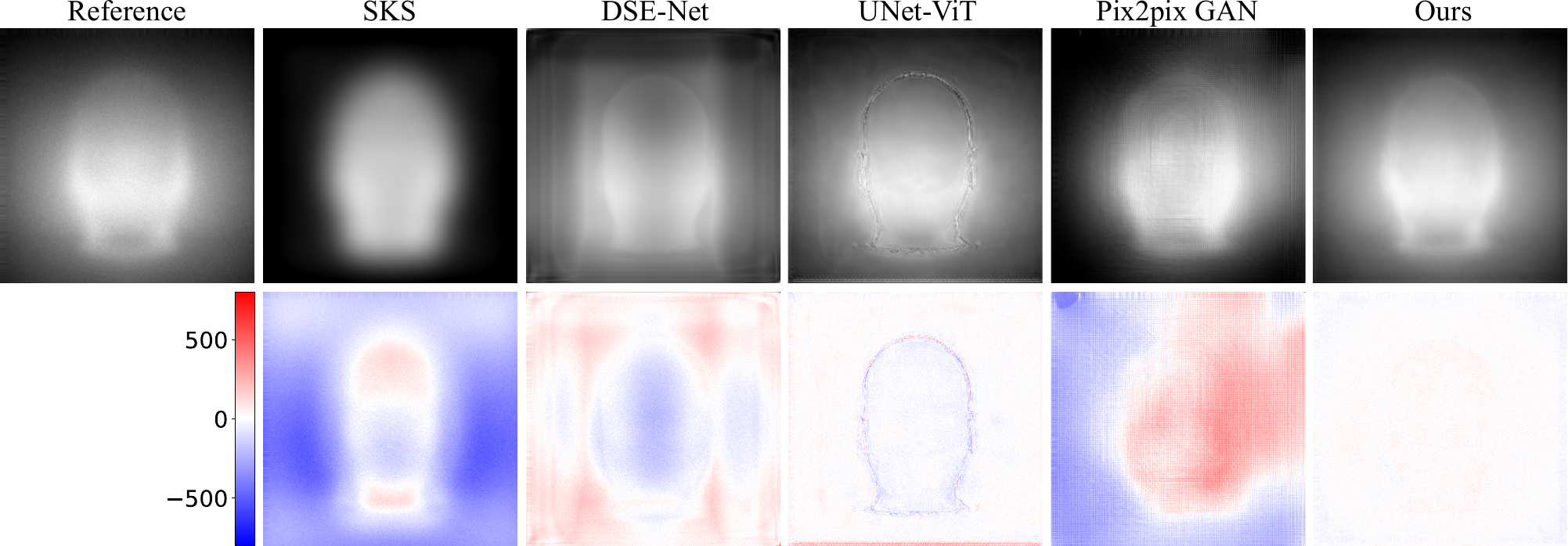} 
    \caption{Synthetic experiment results on the estimated scatter signal using different methods. From left to right are Reference, SKS, DSE-Net, UNet-ViT, Pix2pix GAN, and our method. The first row displays the projection domain scatter images, while the second row shows the difference images relative to the reference image. Display windows: $[300, 1400]$ photons.}
    \label{fig:Fig.5.} 
\end{figure*}

This section presents the results and analysis of synthetic and real-scan experiments. The effectiveness of scatter correction is demonstrated from multiple perspectives, including scatter signal and reconstructed images. In addition, quantitative evaluations are conducted to further assess the performance of the proposed method.

\subsection{Results on Synthetic Data Experiment}

\subsubsection{Estimated Scatter Signal}

We evaluate the estimated scatter outputs using the validation dataset. Fig. \ref{fig:Fig.5.} shows the scatter images at a representative projection angle obtained by different methods: (a) ground truth scatter distribution, (b)–(e) correspond to the results from the SKS method, DSE-Net, UNet-ViT, and Pix2pix GAN, respectively, and (f) shows the result from the proposed method. The reference image reveals that the scatter signal exhibits a low-frequency characteristics and is highly correlated with the head structure at this angle. The SKS method, which estimates scatter using Gaussian kernels, produces results that roughly match the structural layout of the reference. DSE-Net captures the overall distribution of the scatter but demonstrates limited mapping capability. UNet-ViT, which incorporates a Transformer module into the UNet architecture, enhances global feature learning, yet struggles with edge preservation in the head region. Pix2pix GAN successfully captures the general contour of the scatter. Visually, our proposed method achieves the closest match to the reference scatter distribution.

A quantitative analysis was conducted on the predicted scatter results, 
\begin{table}[htbp]
  \centering 
  \caption{Quantitative evaluation (mean ± std) for the estimated scatter signal, as shown in Fig.~\ref{fig:Fig.5.}, based on PSNR, SSIM, and RMSE.}
  \label{tab:denoising_performance}
  \renewcommand{\arraystretch}{1.2}
  \begin{tabular}{lrrr}
    \toprule 
    \textbf{Method} & \textbf{PSNR (dB)} & \textbf{SSIM} & \textbf{RMSE (photons)} \\
    \midrule
    SKS         & $13.53 \pm 0.61$ & $0.62 \pm 0.04$ & $325.05 \pm 57.21$ \\
    DSE-net     & $25.40 \pm 2.63$ & $0.69 \pm 0.05$ & $ 82.84 \pm 13.88$ \\
    Pix2pix GAN & $14.39 \pm 1.58$ & $0.18 \pm 0.04$ & $304.93 \pm 97.61$ \\
    Unet-ViT    & $28.49 \pm 1.84$ & $0.65 \pm 0.06$ & $ 57.17 \pm 11.56$ \\
    Ours & $\mathbf{33.18 \pm 1.34}$ & $\mathbf{0.73 \pm 0.05}$ & $\mathbf{33.36 \pm 2.01}$ \\
    \bottomrule 
  \end{tabular}
\end{table}
as shown in Table \ref{tab:denoising_performance}. Each case in the validation set contains 360 projection images. Accordingly, for each method, the predicted scatter maps were evaluated against the reference images on a per-angle basis to compute the mean and standard deviation of performance metrics. It can be observed that the proposed method consistently outperforms others in terms of PSNR, SSIM, and RMSE. Moreover, it yields the lowest standard deviation, indicating superior stability.

\subsubsection{Corrected Reconstructed Slices}

\begin{figure*}[!t]  
    \centering
    \includegraphics[width=0.95\textwidth]{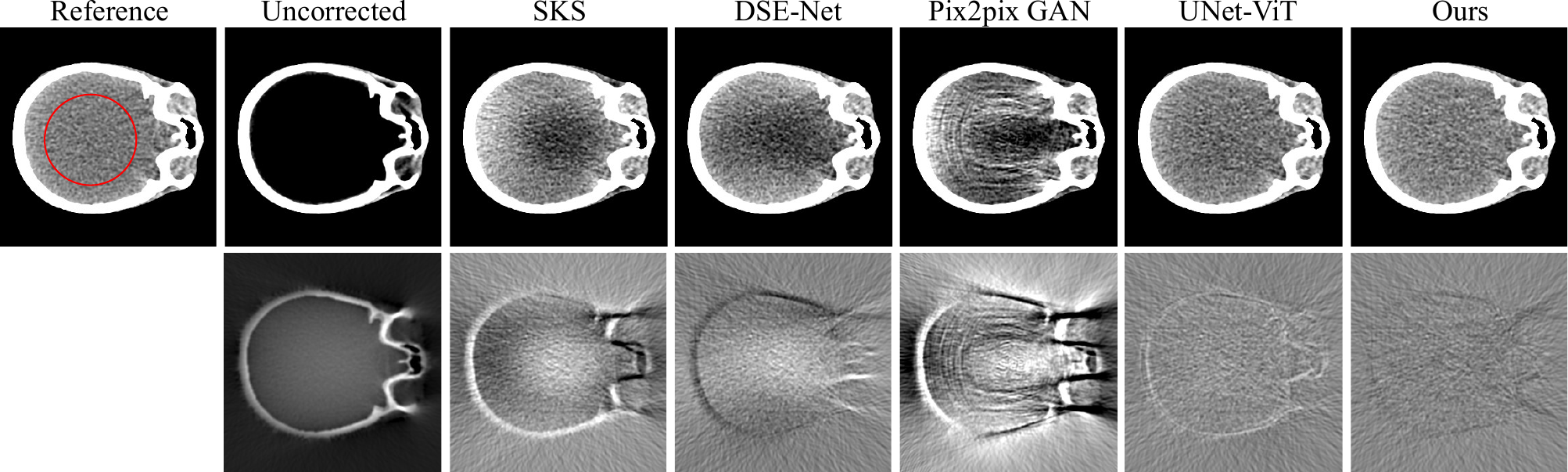} 
    \caption{Synthetic experiment reconstructed results on scatter correction using compared methods. The first row is reconstructed images, while the second row shows the difference images relative to the reference image. The display window of uncorrected error image set [-100, 800] HU. The others display windows are defined as [-80, 80] HU.}
    \label{fig:Fig.8.} 
\end{figure*}

Scatter correction was applied to the measured data, followed by image reconstruction using the FDK algorithm. 
\begin{figure}[h]  
    \centering
    \includegraphics[width=0.45\textwidth]{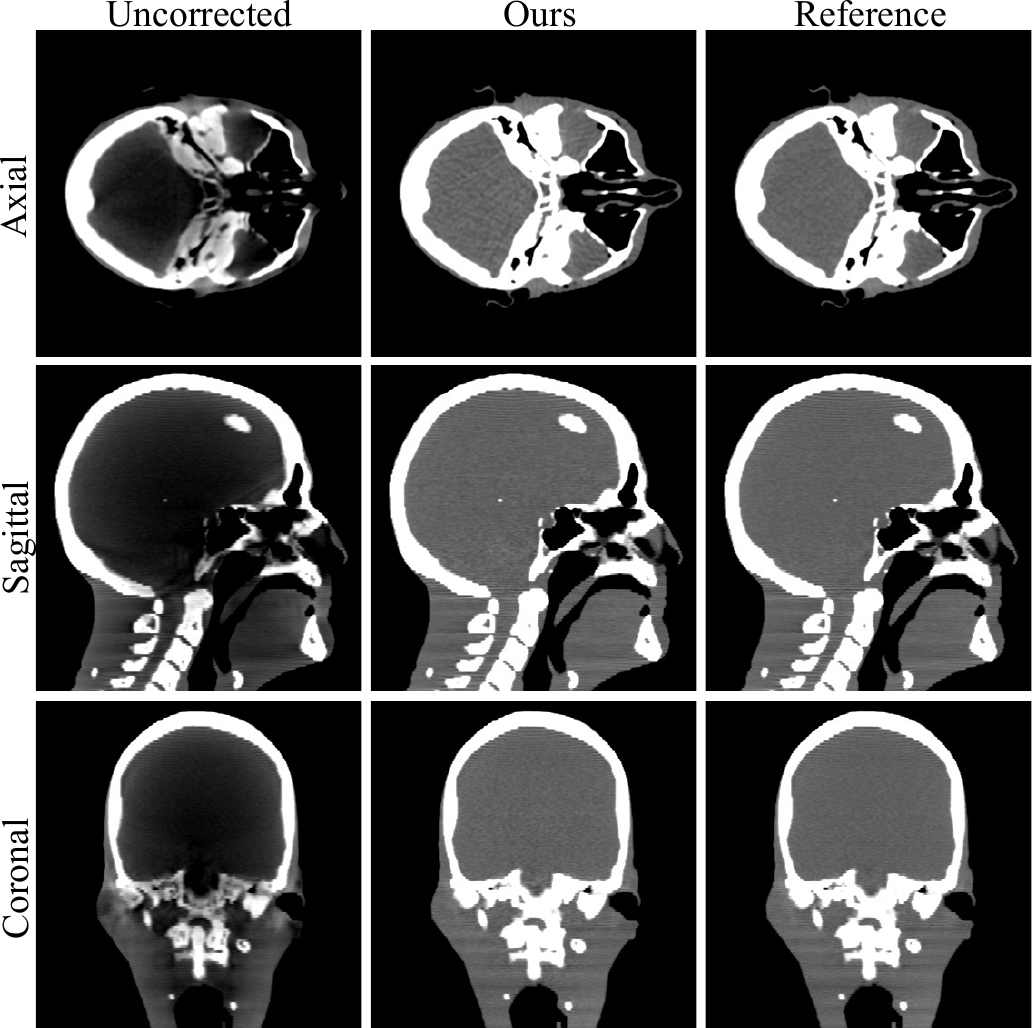} 
    \caption{Synthetic experiment reconstructed results on scatter correction using our proposed method. The axial, sagittal, and coronal views of the head phantom image,the left column is uncorrected image,the middle column is ours and the right column is the reference. Display window: [-300,500] HU.} 
    \label{fig:Fig.7.} 
\end{figure}
Fig. \ref{fig:Fig.7.} presents the results on uncorrected reconstructed slices, corrected reconstructed slices using the proposed method, and the reference reconstructed slices without scatter. It can be observed that, without scatter correction, the reconstructed image exhibits generally lower HU values and reduced contrast. After applying the proposed correction method, the HU values and image contrast are significantly improved. Visually, the result produced by the proposed method is nearly indistinguishable from the scatter-free reference.

\subsubsection{Comparion with other methods}

As shown in Fig. \ref{fig:Fig.8.}. Scatter corrections were performed separately using the comparison methods above, each reconstructed using the FDK algorithm. The first column shows the reconstruction results of the different methods, and the second column shows the error images with the reference results. It can be seen that the effect of scattered photons leads to a decrease in the HU value in the head and a dark area. The scatter artifacts are substantially improved after correction by the different methods. However, the window width $[-80,80]$ HU is a very narrow window width, the comparison methods still suffer from unclean scatter correction, and as can be seen from the error images, the method in this paper has the best scatter correction results.

In order to further verify the scatter correction precision, the values in the red circles in Fig. \ref{fig:Fig.8.} are quantitatively measured, 
\begin{table}[htbp]
  \centering 
  \caption{Synthetic experiment reconstructed slices ROI values (mean ± std) of red circles in Fig.~\ref{fig:Fig.8.}. Error is computed as mean$-$Reference mean (Reference mean = $-1.686$\,HU).}
  \label{tab:hu_roi}
  \renewcommand{\arraystretch}{1.2}
  \begin{tabular}{lrr}
    \toprule 
    \textbf{Method} & \textbf{ROI Value (HU)} & \textbf{Error (HU)} \\
    \midrule
    Reference   & $-1.686 \pm 12.244$   & $/$ \\
    Uncorrected & $-217.010 \pm 40.290$ & $215.324$ \\
    SKS         & $-5.432 \pm 27.986$   & $3.746$ \\
    DSE-net     & $-13.977 \pm 23.563$  & $12.291$ \\
    Pix2pix GAN & $-12.179 \pm 33.872$  & $10.493$ \\
    Unet-ViT    & $-6.059 \pm 15.917$   & $4.373$ \\
    Ours        & $\mathbf{1.204 \pm 14.333}$    & $\mathbf{2.890}$ \\
    \bottomrule 
  \end{tabular}
\end{table}
as shown in Table \ref{tab:hu_roi}. It can be seen that the values corrected by the method of this paper are closest to the reference value, and the noise level is not amplified from the standard deviation.

Table \ref{tab:quantitative_metrics} provides the average 
\begin{table}[htbp]
  \centering 
  \caption{Quantitative evaluation (mean ± standard deviation) for synthetic experiment reconstructed slices of comparison methods using PSNR (dB), SSIM, and RMSE (HU).}
  \label{tab:quantitative_metrics}
  \sisetup{separate-uncertainty = true}
  \begin{tabular}{
    l
    S[table-format=2.2(2)]
    S[table-format=1.3(3)]
    S[table-format=2.2(2)]
  }
    \toprule
    \textbf{Methods} & \textbf{PSNR (dB)} & \textbf{SSIM} & \textbf{RMSE (HU)} \\
    \midrule
    Uncorrected   & 29.46 \pm 3.62 & 0.915 \pm 0.064 & 69.99 \pm 51.42 \\
    SKS           & 43.17 \pm 5.57 & 0.977 \pm 0.018 & 13.01 \pm 12.68 \\
    DSE-Net       & 48.24 \pm 4.49 & 0.986 \pm 0.012 &  7.23 \pm 5.85  \\
    Pix2pix GAN   & 38.58 \pm 5.76 & 0.957 \pm 0.037 & 24.26 \pm 25.27 \\
    UNet-ViT      & 50.89 \pm 4.01 & 0.993 \pm 0.006 &  4.79 \pm 3.23  \\
    Ours          & $\mathbf{52.87 \pm 4.36}$ & $\mathbf{0.995 \pm 0.004}$ &  $\mathbf{4.03 \pm 2.98}$  \\
    \bottomrule
  \end{tabular}
\end{table}
statistical quantitative evaluation of reconstructed slices using compared methods,  including the means and standard deviations of PSNR, SSIM, and RMSE. The size of reconstructed slices is $512\times512\times512$.It is also seen that scatter correction methods present competitive performance compared to uncorrected images. In particular, our proposed method achieves the best scores in all evaluation metrics, further demonstrating its superiority.

\section{Discussion and Conclusion}

This study proposes a physics-inspired scatter correction method for CBCT. To more accurately model the point scatter probability distribution function, we introduce a Gaussian RBF to parameterize the scatter signal distribution and embed it into the KAN modules within a U-shaped backbone. This design enables efficient modeling and fitting of scatter features. The proposed method through synthetic and cross-platform real-scan experiments. In synthetic experiments, the reconstruction error was reduced from 215 HU to 3 HU and the standard deviation indicated improved image uniformity. Compared with several existing methods, it demonstrates superior performance in terms of both visual quality and quantitative metrics, including PSNR, RMSE, and HU deviation.


\end{document}